\documentclass[10pt,conference]{IEEEtran}
\IEEEoverridecommandlockouts
\usepackage{cite}
\usepackage{amsmath,amssymb,amsfonts}
\usepackage{algorithmic}
\usepackage{graphicx}
\usepackage{textcomp}
\usepackage{xcolor}

\usepackage{multirow}
\usepackage{makecell}
\usepackage{adjustbox}
\usepackage{pifont}
\usepackage{subcaption}
\usepackage{rotating}
\usepackage{caption}
\usepackage{array}
\newcolumntype{C}[1]{>{\centering\arraybackslash}p{#1}} %
\usepackage{booktabs}
\usepackage[hidelinks, colorlinks=true, linkcolor=blue]{hyperref}
\newcommand{\corr}{\textsuperscript{\footnotesize *}}
\DeclareRobustCommand{\IEEEauthorrefmark}[1]{\smash{\textsuperscript{\footnotesize #1}}}
\usepackage{xcolor}
\newcommand{\graycross}{\textcolor{gray!50}{\ding{55}}}
\usepackage{float}
\newcommand{\raisedast}{\mathord{\raisebox{0.3ex}{*}}}
\IEEEaftertitletext{\vspace{-12pt}}

\def\BibTeX{{\rm B\kern-.05em{\sc i\kern-.025em b}\kern-.08em
    T\kern-.1667em\lower.7ex\hbox{E}\kern-.125emX}}
\begin{document}

\title{Extending Precipitation Nowcasting Horizons via Spectral Fusion of Radar Observations and Foundation Model Priors}

\author{
    \IEEEauthorblockN{
        Yuze Qin\IEEEauthorrefmark{1}, 
        Qingyong Li\IEEEauthorrefmark{1}, 
        Zhiqing Guo\IEEEauthorrefmark{1},
        Wen Wang\IEEEauthorrefmark{1},
        Yan Liu\IEEEauthorrefmark{1,2},
        Yangli-ao Geng\IEEEauthorrefmark{1,2}\corr\thanks{$\raisedast$ Corresponding author.}
    }
    \IEEEauthorblockA{
        \small
        \IEEEauthorrefmark{1}Key Laboratory of Big Data \& Artificial Intelligence in Transportation (Ministry of Education), Beijing Jiaotong University, Beijing, China\\
    }
    \IEEEauthorblockA{
        \small
        \IEEEauthorrefmark{2}CMA Key Laboratory of Transportation Meteorology, Nanjing Innovation Institute for Atmospheric Sciences, Nanjing, China\\
    }    
    \IEEEauthorblockA{
        \small
        \{qinyuze, liqy, 23111126, wangwen, gengyla\}@bjtu.edu.cn, yanliu@cma.gov.cn
    }
}

\maketitle

\begin{abstract}
Precipitation nowcasting is critical for disaster mitigation and aviation safety. However, radar-only models frequently suffer from a lack of large-scale atmospheric context, leading to performance degradation at longer lead times. While integrating meteorological variables predicted by weather foundation models offers a potential remedy, existing architectures fail to reconcile the profound representational heterogeneities between radar imagery and meteorological data. To bridge this gap, we propose PW-FouCast, a novel frequency-domain fusion framework that leverages Pangu-Weather forecasts as spectral priors within a Fourier-based backbone. Our architecture introduces three key innovations: (\textit{i}) Pangu-Weather-guided Frequency Modulation to align spectral magnitudes and phases with meteorological priors; (\textit{ii}) Frequency Memory to correct phase discrepancies and preserve temporal evolution; and (\textit{iii}) Inverted Frequency Attention to reconstruct high-frequency details typically lost in spectral filtering. Extensive experiments on the SEVIR and MeteoNet benchmarks demonstrate that PW-FouCast achieves state-of-the-art performance, effectively extending the reliable forecast horizon while maintaining structural fidelity. Our code is available at \href{https://github.com/Onemissed/PW-FouCast}{https://github.com/Onemissed/PW-FouCast}.
\end{abstract}


\vspace{-12pt}
\section{Introduction}
\begin{figure}[H]
  \centering
  \includegraphics[width=1.0\columnwidth]
  {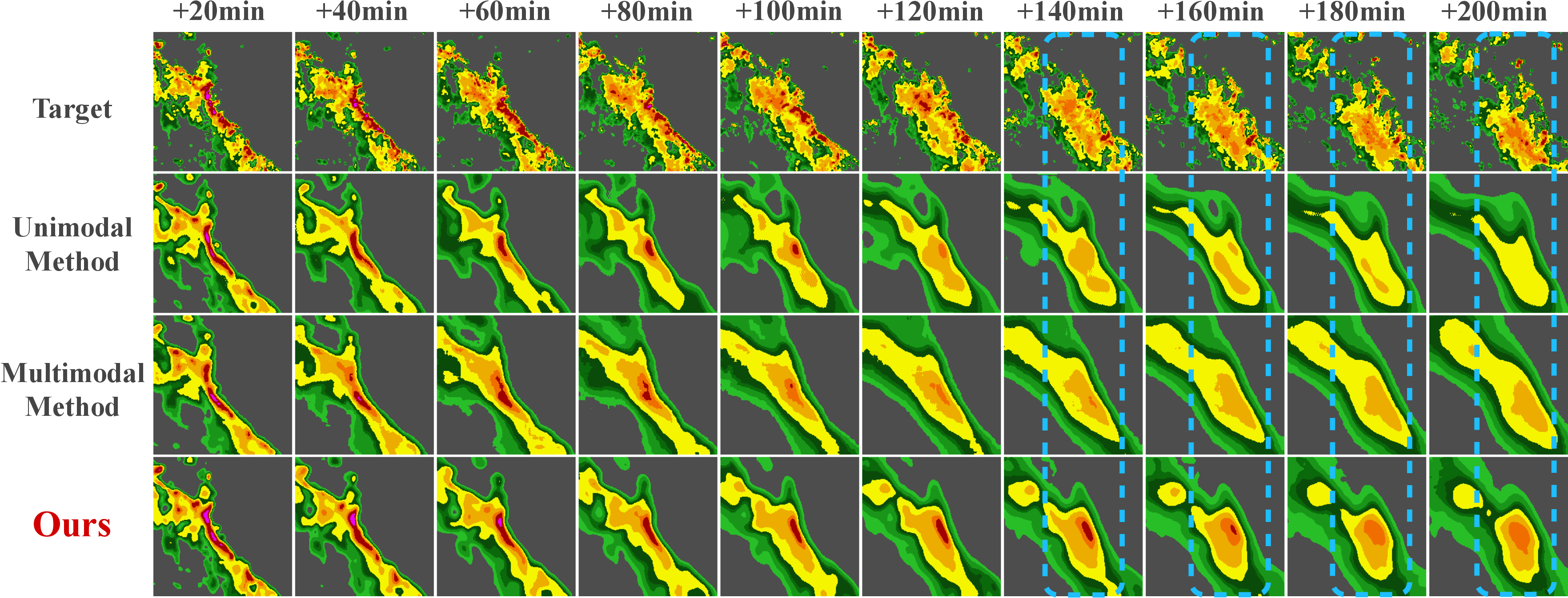}
  \caption{Qualitative comparison of predicted radar reflectivity across unimodal and multimodal methods.}
  \label{fig7}
\end{figure}

Precipitation nowcasting is designed to generate short-term precipitation field predictions, which are critical for time-sensitive applications such as disaster resilience and aviation safety. Modern methodologies increasingly rely on deep learning to capture the complex interplay between convective-scale evolution and larger-scale atmospheric dynamics. However, as illustrated in Fig.~\ref{fig7} (second row), traditional radar-only models often experience performance degradation at longer lead times~\cite{fengperceptually}. This limitation arises because radar reflectivity captures the resulting precipitation field rather than the underlying thermodynamic and dynamic drivers, such as temperature, humidity, wind speed, and pressure, that govern atmospheric evolution. Consequently, disparate atmospheric states may manifest as similar reflectivity patterns, restricting the model's capacity to disambiguate physical causes and accurately project future developments.

To extend the nowcasting horizon, it is essential to incorporate these causal drivers directly into the architecture. Moving
\vspace{-8pt}

\noindent beyond traditional integration of numerical weather prediction (NWP) data~\cite{huang2024tcp}, we utilize weather foundation model outputs as multimodal inputs, specifically because their enhanced predictive precision and computational efficiency provide a more robust basis for improving nowcasting performance. Nevertheless, existing multimodal approaches frequently employ conventional spatial integration schemes such as addition, concatenation or cross-attention~\cite{geng2019lightnet,ma2023mm,zheng2024cross}. This direct fusion fails to address the fundamental heterogeneities inherent in these distinct data sources, including differing spatial scales, magnitudes, and temporal evolution patterns~\cite{yu2025pimmnet}. As illustrated in the third row of Fig.~\ref{fig7}, such methods often fail to fully exploit cross-modal synergies, yielding marginal improvements for long-lead nowcasting.

In this work, we propose PW-FouCast, a Fourier-domain backbone that leverages Pangu-Weather~\cite{bi2023accurate} forecasts to overcome these challenges through spectral integration.  Our framework explicitly aligns and fuses spectral amplitude and phase information in the frequency domain, enabling the model to exploit shared phase representations between radar observations and meteorological forecasts. Furthermore, we develop a Frequency Memory module that stores and retrieves historical spectral patterns to correct phase discrepancies dynamically.

The primary contributions of this work are as follows:
\begin{IEEEenumerate}
\item We present a frequency-domain encoder-decoder framework specifically designed to extend nowcasting horizons by effectively assimilating foundation model priors.
\item We propose a novel method to integrate meteorological forecasts with radar reflectivity in the frequency domain, effectively resolving fundamental heterogeneities between these modalities.
\item We design a specialized Frequency Memory module to store and retrieve spectral features of diverse precipitation patterns, enhancing the model's ability to maintain structural fidelity over time.
\item Extensive experiments on the SEVIR and MeteoNet benchmarks demonstrate that PW-FouCast achieves state-of-the-art results, outperforming both radar-only and standard multi-modal baselines.
\end{IEEEenumerate}

\section{Related Work}
\subsection{Uni-modal Spatial-temporal Forecasting}
Uni-modal spatial-temporal forecasting models typically use radar reflectivity as input and can be categorized into recurrent models and non-recurrent models. Recurrent models generate predictions sequentially, one frame at a time, which makes them effective at modeling short-term dependencies. One of the earliest is ConvLSTM~\cite{shi2015convolutional}, which extends LSTM by replacing internal dense operations with convolutions, enabling the network to capture spatial and temporal dependencies in a unified manner. PredRNN~\cite{wang2017predrnn} extends this idea with a “zigzag” memory that flows across time and depth to exchange spatial–temporal representations, and PredRNN v2~\cite{wang2022predrnn} adds reverse scheduled sampling and a decoupling loss to better learn long-range dependencies. LMC-Memory~\cite{lee2021video} further augments recurrent predictors with an external memory and a two-phase alignment scheme to store and recall long-term motion patterns. Despite these advances, recurrent models can be computationally costly, prone to error accumulation over long horizons, and often learn redundant short-term features.

Non-recurrent models predict all frames simultaneously, are computationally efficient, and can capture global spatiotemporal context. SimVP v2~\cite{tan2025simvpv2} and TAU~\cite{tan2023temporal} are pure CNN architectures that employ large-kernel convolutions in their Translator modules to approximate attention and capture global context. Earthformer~\cite{gao2022earthformer} applies self-attention within non-overlapping spatio-temporal cuboids and propagates global context via learnable vectors. PastNet~\cite{wu2024pastnet} injects spectral inductive biases and discretizes feature vectors with a memory bank. AlphaPre~\cite{lin2025alphapre} decomposes forecasts into Fourier-domain phase and amplitude streams fused by an AlphaMixer. NowcastNet~\cite{zhang2023skilful} predicts motion and intensity residuals to warp frames and then refines them with a generative module. Nonetheless, these non-recurrent designs still have difficulty modeling long-term temporal dependencies and lack flexibility for producing variable-length predictions.

\subsection{Multi-modal Spatial-temporal Forecasting}
Multi-modal spatio-temporal models incorporate auxiliary data, such as satellite imagery and meteorological fields, to improve precipitation forecasting. Examples include LightNet~\cite{geng2019lightnet}, which uses dual spatio–temporal encoders to process multiple sources, MM-RNN~\cite{ma2023mm}, which extracts multiscale features from radar and meteorological streams and fuses them with a cross attention–based module, and CM-STJointNet~\cite{zheng2024cross}, which jointly learns radar extrapolation and satellite (IR) prediction via a STJointNet backbone. However, satellite and meteorological fields differ substantially from radar in scale, distribution, and representation, and many multimodal methods do not explicitly resolve these heterogeneities. By exploiting similar phases, our method instead aligns and fuses radar and meteorological information in the frequency domain, enabling more effective cross-modal integration and improved nowcasting skill.

\begin{figure*}[htbp]
  \centering
  \includegraphics[width=0.76\textwidth]{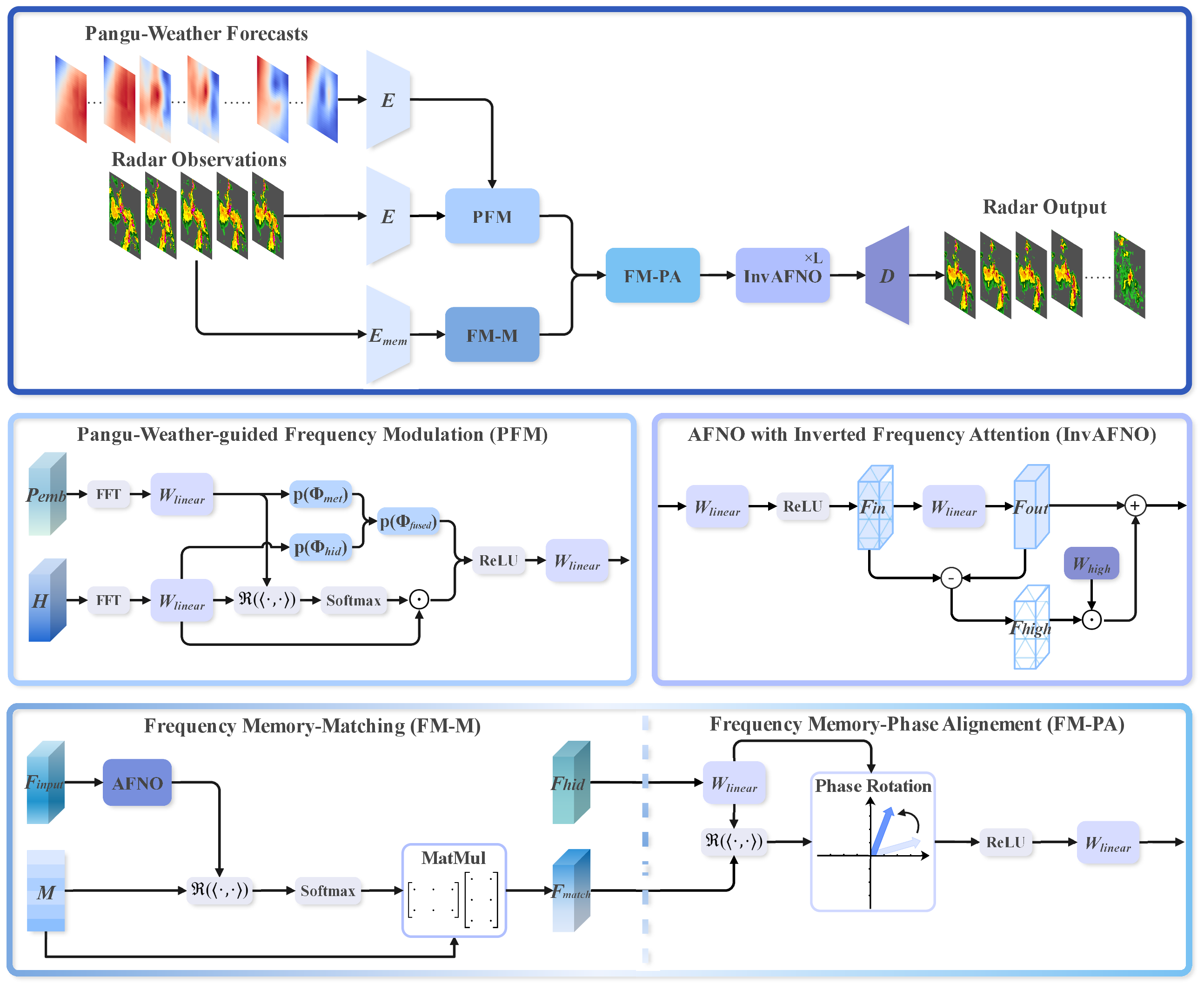}
  \caption{Overall architecture of our proposed PW-FouCast.}
  \label{fig1}
\end{figure*}

\section{Preliminaries}
\subsection{Pangu-Weather Model}
Pangu-Weather is a global weather foundation model trained on 39 years of ERA5~\cite{hersbach2020era5} reanalysis data (1979–2017) at a $0.25^{\circ}$ horizontal resolution. Its 3D Earth-specific transformer (3DEST) architecture captures complex atmospheric dependencies by integrating height as a distinct dimension. The model forecasts five upper-air variables across 13 vertical pressure levels and four surface variables, outperforming the ECMWF’s operational Integrated Forecasting System (IFS) in accuracy. In our framework, we utilize the predicted geopotential, humidity, temperature, and wind components ($u, v$) as multimodal inputs. These variables serve as physical constraints that represent synoptic-scale trends, enabling the model to better maintain structural consistency in long-term precipitation nowcasting.

\subsection{Adaptive Fourier Neural Operator}
The Adaptive Fourier Neural Operator (AFNO)~\cite{guibas2021efficient} introduces an efficient token mixing mechanism in the Fourier domain, extending neural operator frameworks for vision tasks. Given an input feature map $X$, AFNO first applies the forward Fourier transform to perform spatial mixing
\begin{equation}
Z = \mathcal{F}(X),
\end{equation}
where $\mathcal{F}$ denotes the Fourier transform. The channels of resulting spectral features $Z$ are then adaptively mixed using a shared multi-layer perceptron (MLP)
\begin{equation}
\hat{Z}  = \mathrm{MLP}(Z) = W_{2}\,\sigma\left(W_{1}Z\right),
\end{equation}
where $W_1$ and $W_2$ are block-diagonal complex-valued weight matrices, $\sigma$ is the ReLU activation function. All weights are shared across spatial tokens to promote parameter efficiency.

\subsection{Problem Formulation}\label{AA}
We formulate precipitation nowcasting as a spatiotemporal forecasting problem. Let $\{ X_{t} \}_{t=-T+1}^{0}$ be the observed radar sequence of length $T$), where $X_t \in \mathbb{R}^{C \times H \times W}$ denotes a frame with $C$ channels and spatial resolution $H\times W$. We are also given $N$ meteorological forecasts from Pangu-Weather, $\{ P_{i} \}_{i=1}^{N}$, with $P_i \in \mathbb{R}^{M \times H' \times W'}$ containing $M$ variables on an $H'\times W'$ grid at lead time $i$. The task is to predict the future $K$ frames $\{ X_{t} \}_{t=1}^{K}$.

Because the meteorological forecasts and radar observations differ in spatial and temporal resolution, we apply a preprocessing operator $\mathcal{R}(\cdot)$ that spatially regrids and temporally resamples the meteorological fields so they share the radar’s shape and cadence in the model latent space (spatial and temporal interpolation). Denoting the aligned covariates by $\widetilde{P}_{1:K}=\mathcal{R}(P_{1:N})$, our model predicts
\begin{equation}
\hat{X}_{1:K}=f_\theta\big(X_{-T+1:0},,\widetilde{P}_{1:K}\big),
\end{equation}
where $\hat{X}_t\in\mathbb{R}^{C\times H\times W}$ is the predicted radar frame at lead time $t$.

\vspace{-8pt}
\section{Methodology}
\subsection{Overview}
We propose a frequency-domain encoder–decoder architecture that integrates three principal contributions: (\textit{i}) Pangu-Weather–guided Frequency Modulation (PFM), which steers the model’s spectral magnitudes and phase toward the ground truth; (\textit{ii}) Frequency Memory (FM), a learned repository of ground-truth spectral patterns whose memory-matching produces matched frequency features used to correct hidden-layer phases; and (\textit{iii}) Inverted Frequency Attention (IFA), a residual-reinjection mechanism that recovers high-frequency components attenuated by the learned frequency attention. The overall model architecture is illustrated in Fig.~\ref{fig1}.

\subsection{Pangu-Weather-guided Frequency Modulation}
\begin{figure}[H]
  \raggedright
  \includegraphics[width=1.0\columnwidth]{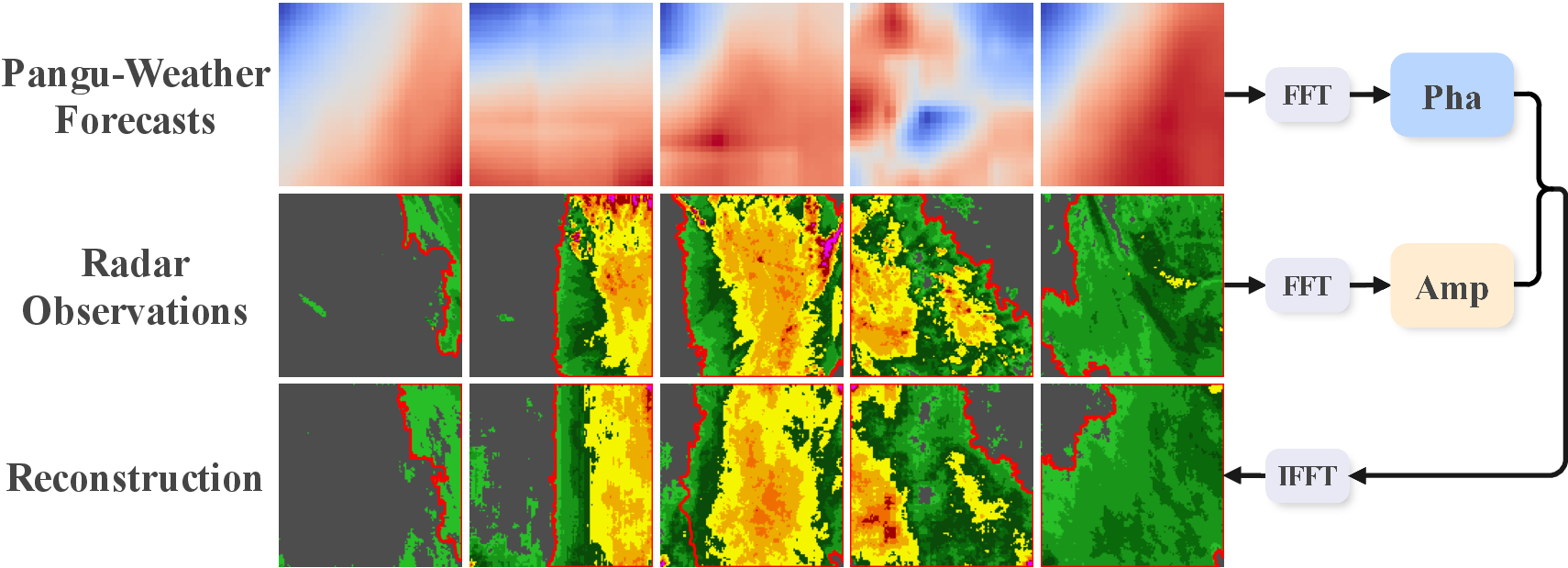}
  \caption{Radar\hspace{0.28em}field\hspace{0.28em}reconstruction\hspace{0.28em}after\hspace{0.28em}integrating\hspace{0.28em}the\hspace{0.28em}amplit\-ude\hspace{0.24em}of\hspace{0.24em}observations\hspace{0.24em}with\hspace{0.24em}the\hspace{0.24em}phase\hspace{0.24em}of\hspace{0.24em}meteorological\hspace{0.24em}variables.}
  \label{fig3}
\end{figure}

As illustrated in Fig.~\ref{fig3}, although Pangu-Weather forecasts differ from radar reflectivity in amplitude and morphology, 

\vspace{-16pt}

\noindent reconstructing a radar-like field by combining the radar amplitude with the phase of Pangu-Weather fields produces a spatial pattern that similarly matches the observed radar reflectivity. This empirical phase similarity indicates that Pangu-Weather forecasts encode useful structural priors, we therefore leverage its phase features to correct and align both the amplitude and phase of the network’s hidden-layer representations.

Concretely, let $F_{\mathrm{hid}}=\mathcal{F}(H) \in \mathbb{C}^{H_{\mathrm{emb}} \times W_{\mathrm{emb}} \times C_{\mathrm{emb}}}$ and $F_{\mathrm{met}}=\mathcal{F}(P_{\mathrm{emb}}) \in \mathbb{C}^{H_{\mathrm{emb}} \times W_{\mathrm{emb}} \times C_{\mathrm{emb}}}$ be the complex Fourier representations of the hidden features and embedded meteorological fields. For each embedding channel $i$ we compute the normalized inner product to quantify phase alignment
\begin{equation}
c_i \;=\; \frac{\langle F_{\mathrm{hid}}^{(i)},\,F_{\mathrm{met}}^{(i)} \rangle}{\lvert F_{\mathrm{hid}}^{(i)}\rvert\;\lvert F_{\mathrm{met}}^{(i)}\rvert},
\end{equation}
take its real part $s_i=\Re(c_i)$ as a scalar similarity score, and convert the scores into channelwise attention maps via a softmax across the $C_{\mathrm{emb}}$ channels
\begin{equation}
w_i \;=\; \frac{\exp(s_i)}{\sum_{j=1}^{C_{\mathrm{emb}}}\exp(s_j)},\qquad i=1,\dots,C_{\mathrm{emb}},
\end{equation}
where $w_i \in \mathbb{R}^{H_{\mathrm{emb}} \times W_{\mathrm{emb}}}$. These attention maps reweight the hidden-feature amplitudes entrywise
\begin{equation}
\hat{A}_{\mathrm{hid}}^{(i)} \;=\; w_i \odot A_{\mathrm{hid}}^{(i)},
\end{equation}
so frequency components whose phase aligns with Pangu-Weather prediction are selectively amplified. We then fuse phases using phasors. Convert a phase angle to a phasor
\begin{equation}
\mathbf{p}(\Phi^{(i)}) = \exp \! \big(\,\mathrm{j}\Phi^{(i)}\big) = \cos\Phi^{(i)} + \mathrm{j}\sin\Phi^{(i)},
\end{equation}
we interpolate between the hidden and meteorological phasors using a learnable parameter $\beta$ and normalize to obtain a unit fused phasor
\begin{equation}
\tilde z^{(i)} = \beta\,\mathbf{p}(\Phi_{\mathrm{hid}}^{(i)}) + (1-\beta)\mathbf{p}(\Phi_{\mathrm{met}}^{(i)}),
\end{equation}
\begin{equation}
\mathbf{p}(\Phi_{\mathrm{fused}}^{(i)}) = \exp \! \big(\,\mathrm{j}\Phi_{\mathrm{fused}}^{(i)}\big) = \frac{\tilde z^{(i)}}{\lvert \tilde z^{(i)}\rvert}.
\end{equation}
Finally, we recombine the amplitude and the fused phasor to form the fused complex coefficients
\begin{equation}
\hat{F}_{\mathrm{hid}}^{(i)} = \hat{A}_{\mathrm{hid}}^{(i)} \cdot \mathbf{p}(\Phi_{\mathrm{fused}}^{(i)}).
\end{equation}
This two-stage procedure first aligns magnitudes according to Pangu-Weather guidance and then refines phases via learned interpolation, yielding hidden-layer frequency coefficients that better match ground truth in both amplitude and phase.

\subsection{Frequency Memory}
Precipitation field variations encompass multiple patterns including movement, expansion, and contraction. The coarse-grained structural priors provided by meteorological variables cannot accurately capture these diverse patterns. Therefore, we design a Frequency Memory module that records phase patterns from observed sequences during training and injects these learned phase features into the prediction pipeline, helping preserve fine-grained structural changes in forecasts.

Following \cite{lee2021video}, training proceeds in two stages. Each stage utilizing two core operations: Frequency Memory-Matching (FM-M) and Frequency Memory-Phase Alignment (FM-PA). In the storing stage (\textit{Phase 1}), the model learns to populate a memory bank $M$ with frequency-domain features derived from ground-truth sequences

Specifically, for the FM-M operation, let $\{ X_{t} \}_{t=1}^{L}$ denote the ground-truth radar sequence and let $E_{\mathrm{mem}}(\cdot)$ be the Frequency Memory encoder. We extract spatio–temporal features and apply a discrete Fourier transform to the encoder output to obtain the ground-truth frequency feature
\begin{equation}
F_{\mathrm{GT}} = \mathcal{F}(E_{\mathrm{mem}}(\{ X_{t} \}_{t=1}^{L})).
\end{equation}
where $F_{\mathrm{GT}} \in \mathbb{C}^{H_{\mathrm{emb}}\times W_{\mathrm{emb}}\times C_{\mathrm{emb}}}$, $H_{\mathrm{emb}},W_{\mathrm{emb}},C_{\mathrm{emb}}$ denote the embedding height, width, and number of channels, respectively. Let $M=\{m_{i}\}_{i=1}^S\in\mathbb{C}^{S\times C_{\mathrm{emb}}}$ denote the Frequency Memory with $S$ slots.  We first normalize $F_{\mathrm{GT}}$ and each memory slot $m_{i}$ elementwise to unit magnitude
\begin{equation}
\hat{F}_{\mathrm{GT}} = \frac{F_{\mathrm{GT}}}{\lvert F_{\mathrm{GT}}\rvert}, \quad \hat{m}_{i} = \frac{m_{i}}{\lvert m_{i}\rvert}
\end{equation}
Using the normalized ground-truth $\hat{F}_{\mathrm{GT}}$ as a complex query, we compute a raw similarity score between the query and each memory slot by taking the real part of their complex inner product
\begin{equation}
\begin{aligned}
s^{\mathrm{raw}}_i(h,w) & = \Re\big(\langle \hat F_{\mathrm{GT}}(h,w,\cdot),\,\hat m_i(\cdot)\rangle\big)
\\ & = \sum_{d=1}^{C_{\mathrm{emb}}}\Big(\Re \big(\hat F_{\mathrm{GT}}^{(d)}(h,w)\big)\Re \big(\hat m_i^{(d)}\big)
+\\ & \hspace{39pt} \Im \big(\hat F_{\mathrm{GT}}^{(d)}(h,w)\big)\Im \big(\hat m_i^{(d)}\big)\Big).
\end{aligned}
\end{equation}
Hence the raw similarity tensor has shape $s^{\mathrm{raw}}\in \mathbb{R}^{H_{\mathrm{emb}}\times W_{\mathrm{emb}}\times S}$, with one $S$-vector of similarities at each spatial location. 

We convert the raw similarities ${s^{\mathrm{raw}}_i(h,w)}_{i=1}^S$ into attention weights by applying a softmax across the memory-slot dimension independently at each spatial location
\begin{equation}
\alpha_i(h,w) = \frac{\exp\!\big(s^{\mathrm{raw}}_i(h,w)\big)}{\sum_{j=1}^{S}\exp\!\big(s^{\mathrm{raw}}_j(h,w)\big)},\qquad i=1,\dots,S.
\end{equation}
Thus $\alpha\in\mathbb{R}^{H_{\mathrm{emb}}\times W_{\mathrm{emb}}\times S}$ is a nonnegative attention map with $\sum_{i=1}^S\alpha_i(h,w)=1$ for every $(h,w)$. The attention maps $\{\alpha_i\}_{i=1}^S$ produce the matched frequency-domain feature via an attention-weighted sum
\begin{equation}
F_{\mathrm{match}}(h,w,d,r)=\sum_{i=1}^{S}\alpha(h,w,i)\;M(i,d,r),
\end{equation}
where $h=1,\dots,H_{\mathrm{emb}}$, $w=1,\dots,W_{\mathrm{emb}}$, $d=1,\dots,C_{\mathrm{emb}}$, and $r\in{1,2}$ indexes real/imag. 
By treating $F_{\mathrm{match}}$ as a convex combination of $M$, the amplitude of $F_{\mathrm{match}}$ is bounded in $[0,1]$, i.e. $\left | F_{\mathrm{match}}\right| \in [0,1]$.

Subsequently, the FM-PA operation leverages the matched frequency feature $F_{\mathrm{match}}$ to correct phase discrepancies within the model's hidden layers. We compute a real-valued raw similarity between $F_{\mathrm{match}}$ and the normalized hidden frequency features. Notably, the amplitude of $F_{\mathrm{match}}$ is not normalized, as it preserves critical information of the recalled spectral patterns.

\begin{equation}
\mathrm{sim} \;=\; \Re \left (\left\langle \frac{\hat{F}_{\mathrm{hid}}}{\lvert \hat{F}_{\mathrm{hid}}\rvert},\,F_{\mathrm{match}}\right\rangle \right ) .
\end{equation}
Because $\mathrm{sim} = \left | F_{\mathrm{match}}\right|\,\mathrm{cos}(\hat{\Phi}_{\mathrm{hid}}-\Phi_{\mathrm{match}}) \in [-1, 1]$, we convert $\mathrm{sim}$ into a phase-fusion weight bounded in $[0,1]$ by
\begin{equation}
w_{\mathrm{phase}} = \tfrac{1}{2}\big(1 - \mathrm{sim}\big).
\end{equation}
As $\mathrm{sim}$ decreases with increasing phase discrepancy, the phase-fusion weight $w_{\mathrm{phase}}$ increases, ensuring that hidden-layer phases are more aggressively aligned with the retrieved spectral patterns. Let the phase difference between $\hat{\Phi}_{\mathrm{hid}}$ and $\Phi_{\mathrm{match}}$ be $\Delta_{\Phi} = \Phi_{\mathrm{match}} - \hat{\Phi}_{\mathrm{hid}}$. 
We rotate the hidden feature phase toward the matched phase by the fraction $w_{\mathrm{phase}}$ of the full phase difference
\begin{equation}
\widetilde{F}_{\mathrm{hid}} = \hat{F}_{\mathrm{hid}} \cdot \exp \! \big(\mathrm{j}\,w_{\mathrm{phase}}\,\Delta_{\Phi}\big).
\end{equation}
This yields the phase-corrected hidden representation $\widetilde{F}_{\mathrm{hid}}$.

At the matching stage (\textit{Phase 2}) we extract features from the input sequence $\{ X_{t} \}_{t=-T+1}^{0}$ and transform the encoder output to the frequency domain
\begin{equation}
F_{\mathrm{input}} = \mathcal{F}(E_{\mathrm{mem}}(\{ X_{t} \}_{t=-T+1}^{0})).
\end{equation}
We then align the channel dimensionality of $F_{\mathrm{input}}$ with the frequency memory $M$ using a AFNO block
\begin{equation}
\hat{F}_{\mathrm{input}} = W_{\mathrm{align}2}\,\sigma\left(W_{\mathrm{align}1}F_{\mathrm{input}}\right),
\end{equation}
where $W_{\mathrm{align}1}$ and $W_{\mathrm{align}2}$ are block-diagonal, complex-valued weight matrices and $\sigma$ denotes the elementwise ReLU activation. The aligned feature $\hat{F}_{\mathrm{input}}$ therefore has shape $\hat{F}_{\mathrm{input}}\in\mathbb{C}^{H_{\mathrm{emb}}\times W_{\mathrm{emb}}\times C_{\mathrm{emb}}}$.

Next, we apply the same memory-matching procedure used for $F_{\mathrm{gt}}$ to $\hat{F}_{\mathrm{input}}$ to obtain the matched frequency features from the frequency memory $M$, and use these matched features to correct the phase of the hidden features. Importantly, the frequency memory $M$ is fixed during phase 2 and is not updated.

\subsection{Inverted Frequency Attention}
The proposed Inverted Frequency Attention is implemented within the hidden layer module to enhance spectral diversity specifically for the extraction of temporal features. The standard frequency attention mechanism utilized for temporal modeling typically takes the following form
\begin{equation}
F_{\text{out}} = W_{\text{learned}} \cdot F_{\text{in}},
\end{equation}
where  $F_{\text{in}} = \mathcal{F}(X_{\text{in}})\in\mathbb{C}^{H\times W\times C}$ denotes the complex Fourier coefficients of the input features and $W_{\text{learned}}\in\mathbb{C}^{H\times W\times C}$ denotes a learnable complex linear operator applied per frequency. 

Empirically, $W_{\text{learned}}$ tends to attenuate small-amplitude coefficients in $F_{\text{in}}$, producing an effective low-pass behaviour similar to the suppression of high-frequency components previously reported for attention-like transforms such as ViT \cite{chen2025frequency}.

Motivated by this observation, we obtain the discarded high-frequency residual by subtracting $F_{\text{out}}$ from $F_{\text{in}}$, effectively applying the inverse mask of $W_{\text{learned}}$
\begin{equation}
F_{\text{high}} = F_{\text{in}} - F_{\text{out}}.
\end{equation}
We then reintroduce high-frequency detail in a controlled manner using a learnable gating vector $w_{\text{high}}\in\mathbb{R}^{1\times 1\times C}$. The gated residual is added back to the low-frequency component
\begin{equation}
\hat{F}_{\text{out}} = F_{\text{out}} + w_{\text{high}} \odot F_{\text{high}},
\end{equation}
where $w_{\text{high}}$ is broadcast across the spatial frequency dimensions during the elementwise multiplication. This achieving the fusion of high-frequency and low-frequency feature in a simple way.

\subsection{Loss Function}
The training objective is a weighted sum of a spatial mean-squared error and a spectral $L_1$ loss
\begin{equation}
\mathcal{L} = \mathbb{E}\left[\left \|X_{1:K} - \hat{X}_{1:K} \right \|_2^2\right] + \lambda\,\mathbb{E}\left[\left \| \mathcal{F}(X_{1:K})- \mathcal{F}(\hat{X}_{1:K}) \right \|_1\right],
\end{equation}
where $0 \leq \lambda \leq 1$ is a hyperparameter. The MSE term penalizes spatial reconstruction error, while the spectral $L_1$ term encourages accurate recovery in frequency space, together they reduce spatial error and help preserve high-frequency echo structure in the predictions.

\begin{table*}[htbp]
  \centering
  \scriptsize
  \setlength{\tabcolsep}{2pt}
  \caption{Quantitative evaluation on the SEVIR dataset. {\bfseries Bold}: best; {\underline{Underline}}: second-best.}\label{tab1}
  \makebox[\textwidth][c]{
{\renewcommand{\arraystretch}{1.1}
  \begin{tabular}{c|c|c c c c c c c|C{0.7cm}|C{0.9cm}|C{0.9cm}|C{0.9cm}|C{0.9cm}}
    \toprule
    \multirow{2}{*}{Type} & \multirow{2}{*}{Model} & \multicolumn{7}{c|}{CSI$\uparrow$} & \multicolumn{1}{c|}{HSS$\uparrow$} & \multirow{2}{*}{MSE$\downarrow$} & \multirow{2}{*}{MAE$\downarrow$} & \multirow{2}{*}{PSNR$\uparrow$} & \multirow{2}{*}{SSIM$\uparrow$} \\
    & & 16 & 74 & 133 & 160 & 181 & 219 & Avg & Avg & & & & \\
    \midrule
    \multirow{9}{*}{Unimodal} & PredRNN v2~\cite{wang2022predrnn} & {\underline{0.5922}} & 0.4757 & 0.2071 & 0.1179 & 0.0879 & 0.0371 & 0.2530 & 0.3376 & {\underline{692.4151}} & {\underline{12.8557}} & 23.5778 & {\underline{0.5675}} \\
    & SimVP v2~\cite{tan2025simvpv2} & 0.5747 & 0.4319 & 0.1806 & 0.0982 & 0.0695 & 0.0299 & 0.2308 & 0.3085 & 735.3561 & 13.3449 & 23.7060 & 0.5488 \\
    & TAU~\cite{tan2023temporal} & 0.5767 & 0.4750 & 0.2328 & 0.1223 & 0.0834 & 0.0376 & 0.2546 & 0.3388 & 739.1168 & 13.5212 & 23.5635 & 0.5503 \\
    & Earthformer~\cite{gao2022earthformer} & 0.5824 & {\underline{0.4846}} & {\underline{0.2346}} & {\underline{0.1300}} & 0.0951 & 0.0441 & 0.2618 & 0.3489 & 717.5891 & 13.5044 & {\underline{23.7495}} & 0.5459 \\
    & PastNet~\cite{wu2024pastnet} & 0.5551 & 0.4616 & 0.2044 & 0.1114 & 0.0800 & 0.0419 & 0.2424 & 0.3236 & 718.3476 & 14.9068 & 22.7698 & 0.3905 \\
    & AlphaPre~\cite{lin2025alphapre} & 0.5737 & 0.4739 & 0.2266 & 0.1176 & 0.0790 & 0.0349 & 0.2510 & 0.3335 & 744.8281 & 13.7112 & 23.6576 & 0.5286 \\
    & NowcastNet~\cite{zhang2023skilful} & 0.5803 & 0.4642 & 0.2200 & 0.1234 & 0.0911 & 0.0444 & 0.2539 & 0.3402 & 750.2669 & 13.3689 & 23.6842 & 0.5595 \\
    & LMC-Memory~\cite{lee2021video} & 0.5643 & 0.4586 & 0.1912 & 0.0997 & 0.0728 & 0.0305 & 0.2362 & 0.3136 & 744.8522 & 13.7543 & 23.5571 & 0.5445 \\
    & AFNO~\cite{guibas2021efficient} & 0.5858 & 0.4804 & 0.2311 & 0.1286 & {\underline{0.0983}} & {\underline{0.0469}} & {\underline{0.2618}} & {\underline{0.3502}} & 740.2969 & 13.2170 & 23.7374 & 0.5576 \\
    \midrule
    \multirow{4}{*}{Multimodal} & LightNet~\cite{geng2019lightnet} & 0.5697 & 0.4627 & 0.1974 & 0.0918 & 0.0574 & 0.0169 & 0.2326 & 0.3053 & 725.6782 & 13.8670 & 23.5267 & 0.5220 \\
    & MM-RNN~\cite{ma2023mm} & 0.5679 & 0.4582 & 0.2052 & 0.0945 & 0.0635 & 0.0231 & 0.2354 & 0.3112 & 750.6046 & 13.7483 & 23.4178 & 0.5374 \\
    & CM-STjointNet~\cite{zheng2024cross} & 0.5825 & 0.4778 & 0.2188 & 0.1218 & 0.0894 & 0.0468 & 0.2562 & 0.3420 & 715.5997 & 13.3922 & 23.6520 & 0.5407 \\
    & Ours & {\bfseries 0.6023} & {\bfseries 0.4900} & {\bfseries 0.2558} & {\bfseries 0.1511} & {\bfseries 0.1163} & {\bfseries 0.0628} & {\bfseries 0.2797} & {\bfseries 0.3757} & {\bfseries 676.6416} & {\bfseries 12.5787} & {\bfseries 24.1511} & {\bfseries 0.5789} \\
    \bottomrule
  \end{tabular}
  }
  }
\end{table*}

\begin{table*}[htbp]
  \centering
  \scriptsize
  \setlength{\tabcolsep}{2pt}
  \caption{Quantitative evaluation on the MeteoNet dataset. {\bfseries Bold}: best; {\underline{Underline}}: second-best.}\label{tab2}
  \makebox[\textwidth][c]{
{\renewcommand{\arraystretch}{1.1}
  \begin{tabular}{c|c|c c c c|c c c c|C{0.9cm}|C{0.9cm}|C{0.9cm}|C{0.9cm}}
    \toprule
    \multirow{2}{*}{Type} & \multirow{2}{*}{Model} & \multicolumn{4}{c|}{CSI$\uparrow$} & \multicolumn{4}{c|}{HSS$\uparrow$} & \multirow{2}{*}{MSE$\downarrow$} & \multirow{2}{*}{MAE$\downarrow$} & \multirow{2}{*}{PSNR$\uparrow$} & \multirow{2}{*}{SSIM$\uparrow$} \\
    & & 12 & 24 & 32 & Avg & 12 & 24 & 32 & Avg & & & & \\
    \midrule
    \multirow{9}{*}{Unimodal} & PredRNN v2~\cite{wang2022predrnn} & 0.3344 & 0.1554 & 0.0350 & 0.1749 & 0.4897 & 0.2647 & 0.0671 & 0.2738 & 9.3032 & 0.7996 & 34.0562 & 0.8501 \\
    & SimVP v2~\cite{tan2025simvpv2} & 0.3250 & 0.1400 & 0.0195 & 0.1615 & 0.4786 & 0.2415 & 0.0379 & 0.2527 & 9.4314 & 0.8343 & 34.9568 & 0.8499 \\
    & TAU~\cite{tan2023temporal} & 0.3444 & 0.1413 & 0.0248 & 0.1702 & 0.5013 & 0.2446 & 0.0482 & 0.2647 & 8.3152 & 0.7907 & 34.8399 & 0.8474 \\
    & Earthformer~\cite{gao2022earthformer} & 0.3628 & 0.1839 & 0.0573 & 0.2013 & 0.5218 & 0.3072 & 0.1079 & 0.3123 & 7.9094 & 0.7642 & 35.4281 & 0.8572 \\
    & PastNet~\cite{wu2024pastnet} & 0.3526 & 0.1644 & 0.0268 & 0.1813 & 0.5110 & 0.2789 & 0.0518 & 0.2806 & 8.1057 & 0.9666 & 34.6571 & 0.7681 \\
    & AlphaPre~\cite{lin2025alphapre} & 0.3722 & 0.1854 & 0.0575 & 0.2050 & 0.5326 & 0.3098 & 0.1083 & 0.3169 & 7.6863 & 1.0026 & 34.7310 & 0.7636 \\
    & NowcastNet~\cite{zhang2023skilful} & 0.3414 & 0.1595 & 0.0660 & 0.1890 & 0.4990 & 0.2722 & 0.1233 & 0.2982 & 7.8276 & 0.7268 & 35.5236 & 0.8651 \\
    & LMC-Memory~\cite{lee2021video} & 0.3505 & 0.1659 & 0.0448 & 0.1871 & 0.5092 & 0.2817 & 0.0854 & 0.2921 & 7.8226 & 0.7675 & 35.2575 & 0.8496 \\
    & AFNO~\cite{guibas2021efficient} & {\underline{0.3739}} & 0.2018 & {\underline{0.0859}} & {\underline{0.2205}} & {\underline{0.5343}} & 0.3325 & {\underline{0.1576}} & {\underline{0.3415}} & {\underline{7.5631}} & 0.9341 & 34.9673 & 0.7864 \\
    \midrule
    \multirow{4}{*}{Multimodal} & LightNet~\cite{geng2019lightnet} & 0.3539 & 0.1649 & 0.0503 & 0.1897 & 0.5128 & 0.2801 & 0.0955 & 0.2961 & 7.6877 & {\underline{0.6864}} & {\underline{35.5315}} & 0.8716 \\
    & MM-RNN~\cite{ma2023mm} & 0.3456 & {\underline{0.2187}} & 0.0519 & 0.2054 & 0.5015 & {\underline{0.3539}} & 0.0981 & 0.3178 & 9.6015 & 0.7580 & 35.2290 & {\underline{0.8737}} \\
    & CM-STjointNet~\cite{zheng2024cross} & 0.3285 & 0.1553 & 0.0542 & 0.1793 & 0.4832 & 0.2656 & 0.1022 & 0.2837 & 8.8143 & 0.9429 & 34.3279 & 0.7974 \\
    & Ours & {\bfseries 0.3744} & {\bfseries 0.2206} & {\bfseries 0.1022} & {\bfseries 0.2324} & {\bfseries 0.5353} & {\bfseries 0.3579} & {\bfseries 0.1848} & {\bfseries 0.3593} & {\bfseries 7.3844} & {\bfseries 0.6603} & {\bfseries 35.8116} & {\bfseries 0.8740} \\
    \bottomrule
  \end{tabular}
  }
  }
\end{table*}

\section{Experiments}
\subsection{Experimental Setup}
\subsubsection{Dataset}
\textbf{SEVIR}~\cite{veillette2020sevir} is a benchmark precipitation forecasting dataset containing 20,393 meteorological events.  For our experiments we selected events from 2018–2019. The period January 2018–May 2019 is used for training and June–November 2019 for testing, yielding 10,776 training samples and 4,053 test samples. Following~\cite{zhang2023skilful}, all models receive 5 input frames (50 minutes) and predict the next 20 frames (200 minutes).

\textbf{MeteoNet}~\cite{Larvor2020MeteoNet} is an open dataset curated by Météo-France that spans 2016–2018 and covers a $550\times 550$ km region in north-western France. For our experiments we construct a 2018 subset: January–August form the training set (5,381 samples) and September–October the test set (1,027 samples). The model receives 5 input frames (50 minutes) and predicts the next 20 frames (200 minutes).

\textbf{Meteorological Variables} are inferenced from the pretrained Pangu-Weather model. We use five upper-air variables at 500, 600, 700 and 850 hPa, crop each forecast to the latitude–longitude bounding box of the corresponding SEVIR or MeteoNet scene, linearly interpolate the fields in time to align with radar timesteps, and spatially resample them to the model hidden-layer resolution $32\times32$. Finally, these variables are concatenated along the channel dimension and standardized via channel-wise $z$-score normalization to serve as model inputs.

\subsubsection{Training Details}
We train all models using the AdamW optimizer with a learning rate of 0.001. The architecture consists of four convolutional encoder-decoder modules, with a hidden layer depth (L) of 6. Radar inputs are linearly normalized to the range $[0, 1]$, and radar reflectivity is interpolated to a size of $128\times128$ pixels. All experiments were conducted on two RTX 3090 GPUs.

\subsubsection{Evaluation Metrics}
We evaluate nowcasting performance using the Critical Success Index (CSI) and Heidke Skill Score (HSS) at multiple thresholds. CSI quantifies event-based accuracy for exceedance of a reflectivity threshold, while HSS measures overall forecast skill relative to random chance. Following prior work~\cite{gao2023prediff,yu2024diffcast}, thresholds for SEVIR are 16, 74, 133, 160, 181, and 219, and for MeteoNet are 12, 24, and 32. For pixel-level continuous accuracy we report Mean Squared Error (MSE) and Mean Absolute Error (MAE); for perceptual assessment we report Peak Signal-to-Noise Ratio (PSNR) and the Structural Similarity Index (SSIM).

\subsubsection{Hyperparameter Selection}
\begin{figure}[H]
\begin{minipage}[b]{1.0\linewidth}
  \centering
  \centerline{\includegraphics[width=2.6cm]{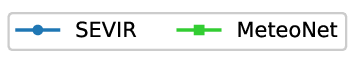}}
\end{minipage}
\begin{minipage}[b]{0.5\linewidth}
  \centering
  \centerline{\includegraphics[width=4.4cm]{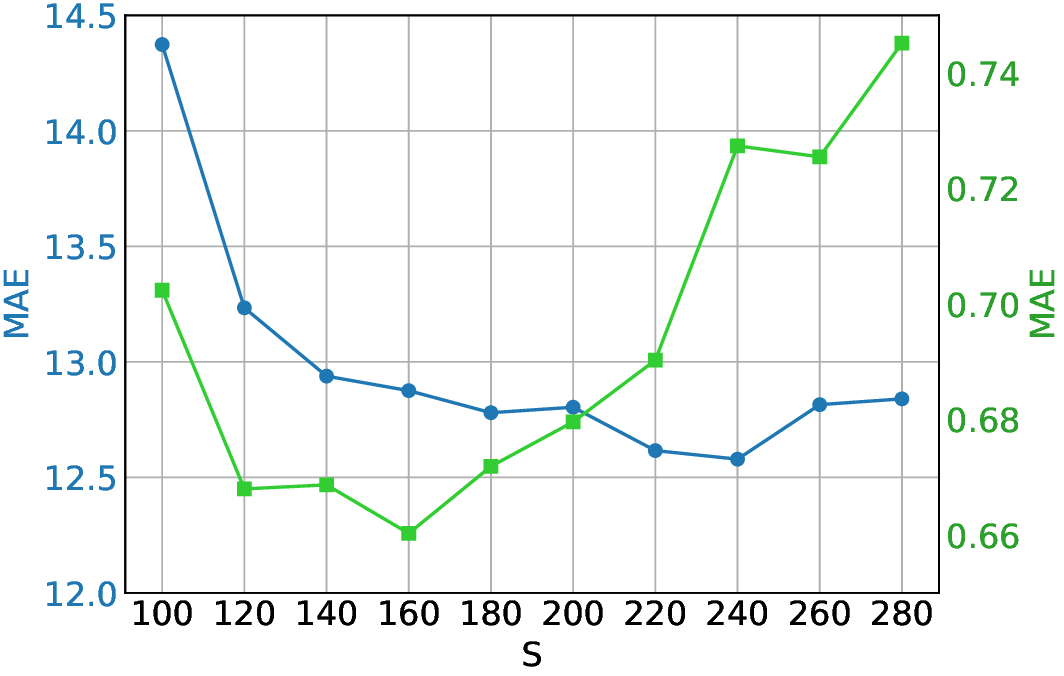}}
\end{minipage}%
\begin{minipage}[b]{0.5\linewidth}
  \centering
  \centerline{\includegraphics[width=4.4cm]{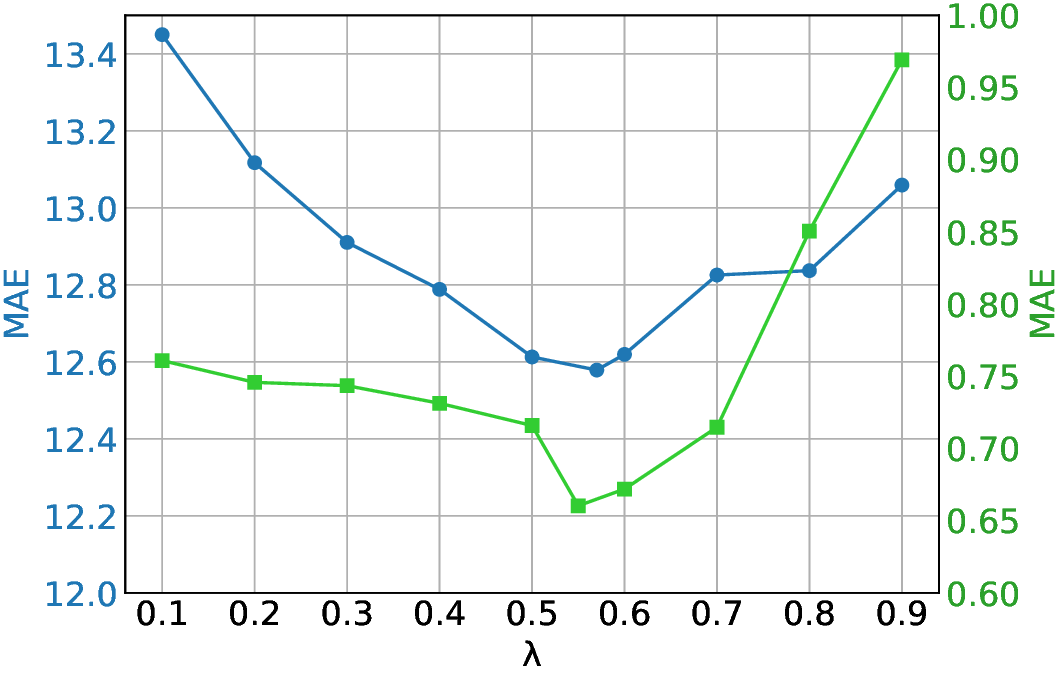}}
\end{minipage}

\caption{Impact of memory slots ($S$) and loss weight ($\lambda$) on model performance (MAE).}
\label{fig2}
\end{figure}

For the number of frequency memory slots $S$ and the loss weight $\lambda$, we performed hyperparameter sweeps evaluated by MAE on the SEVIR and MeteoNet datasets. As shown in Fig.~\ref{fig2}, the sweeps identify optimal memory sizes of $S=240$ for SEVIR and $S=160$ for MeteoNet, a difference we attribute to the greater complexity of precipitation patterns in SEVIR that require more memory capacity. The best loss weights are $\lambda=0.57$ (SEVIR) and $\lambda=0.55$ (MeteoNet), showing consistent behavior across both datasets.

\subsubsection{Baselines}
We compare our approach to twelve state-of-the-art spatiotemporal forecasting models: nine unimodal models PredRNN v2~\cite{wang2022predrnn}, SimVP v2~\cite{tan2025simvpv2}, TAU~\cite{tan2023temporal}, Earthformer~\cite{gao2022earthformer}, PastNet~\cite{wu2024pastnet}, AlphaPre~\cite{lin2025alphapre}, NowcastNet~\cite{zhang2023skilful}, LMC-Memory~\cite{lee2021video} and AFNO~\cite{guibas2021efficient}; and three multimodal models LightNet~\cite{geng2019lightnet}, MM-RNN~\cite{ma2023mm} and CM-STjointNet~\cite{zheng2024cross}.

\subsection{Experimental Results}
As shown in Table~\ref{tab1}, PW-FouCast achieves state-of-the-art performance on the SEVIR dataset, reducing MSE and MAE

\vspace{-6pt}

\noindent by $2.28\%$ and $2.15\%$ while increasing average CSI and HSS by $6.84\%$ and $7.28\%$ over the strongest baselines. On MeteoNet (Table~\ref{tab2}), our model similarly outperforms competitors, reducing MSE and MAE by $2.36\%$ and $3.80\%$, with CSI and HSS improvements of $5.40\%$ and $5.21\%$. Peak PSNR and SSIM scores further demonstrate superior pixel-level accuracy and structural fidelity. These gains confirm that integrating Pangu-Weather spectral priors effectively mitigates the long-lead degradation typical of radar-only models.

Notably, multimodal models like MM-RNN often underperform unimodal baselines because simplistic spatial fusion (e.g., addition or cross-attention) fails to reconcile radar and meteorological heterogeneities. In contrast, PW-FouCast utilizes spectral fusion to align magnitudes and phases with foundation model priors, resolving cross-modal discrepancies that spatial methods cannot adequately address.

\begin{figure}[htb]
\hspace{0.34cm}
\begin{minipage}[b]{0.96\linewidth}
  \centering
  \centerline{\includegraphics[width=8.14cm]{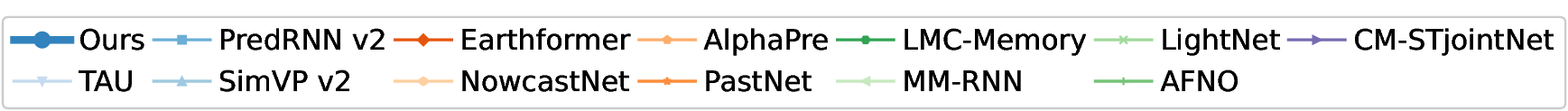}}
\end{minipage}
\begin{minipage}[b]{0.5\linewidth}
  \centering
  \centerline{\includegraphics[width=4.2cm]{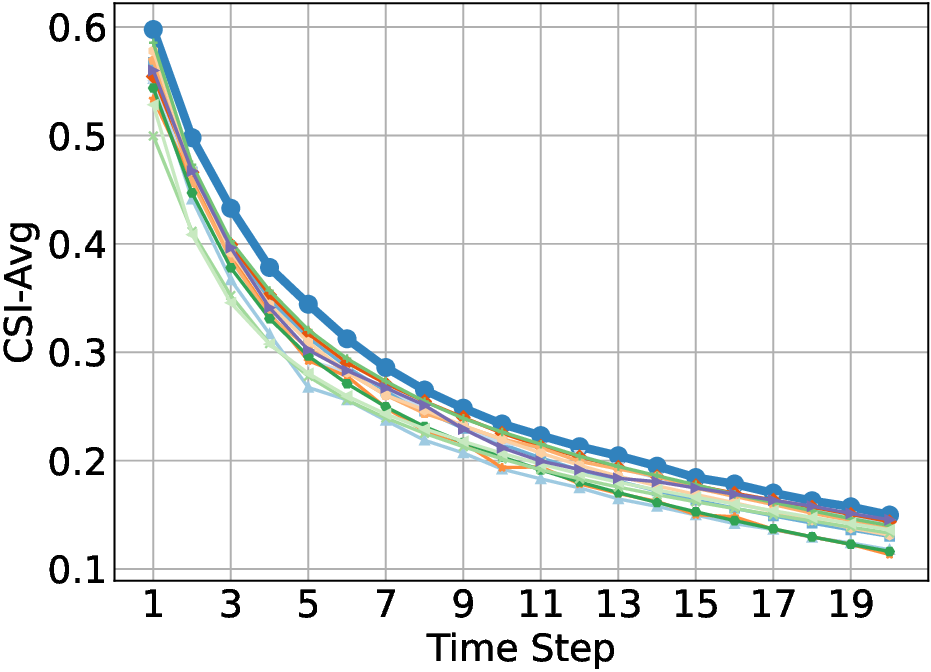}}
\end{minipage}%
\begin{minipage}[b]{0.5\linewidth}
  \centering
  \centerline{\includegraphics[width=4.2cm]{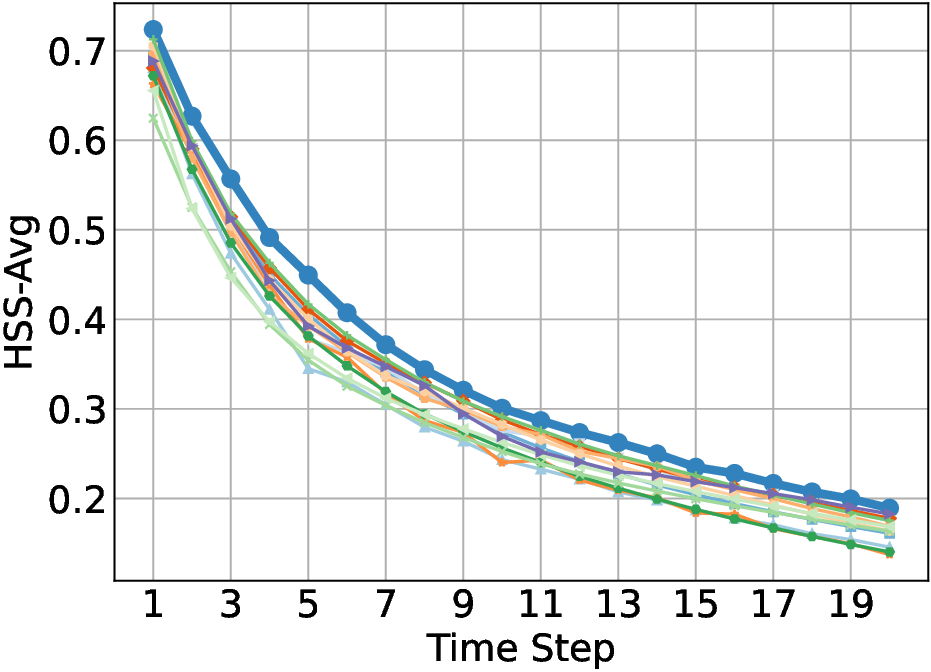}}
\end{minipage}
\begin{minipage}[b]{0.5\linewidth}
  \centering
  \centerline{\includegraphics[width=4.2cm]{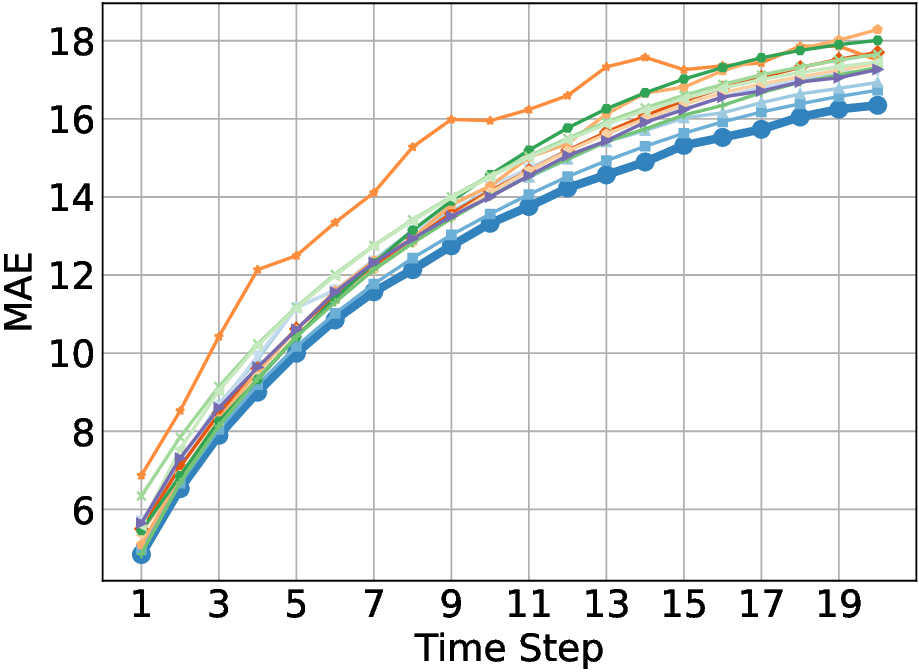}}
\end{minipage}%
\begin{minipage}[b]{0.5\linewidth}
  \centering
  \centerline{\includegraphics[width=4.2cm]{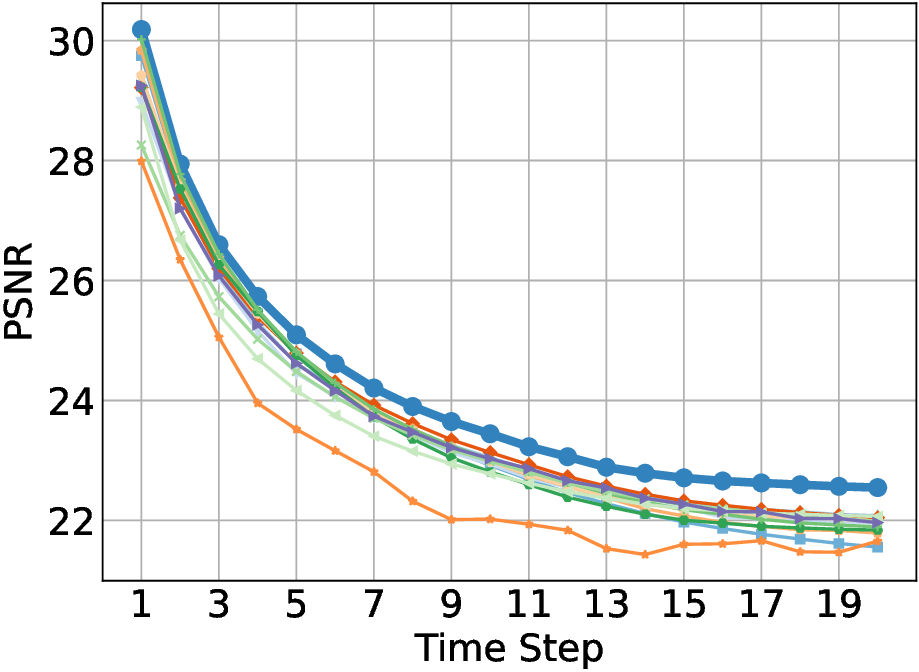}}
\end{minipage}
\caption{Long-term sequence predictive performance on the SEVIR datasets.}
\label{fig4}
\end{figure}

The long-term sequence analysis illustrated in Fig.~\ref{fig4} further confirms that our model maintains a consistent performance lead over all baselines at every individual time step. This is particularly evident in the MAE and PSNR curves, where the performance gap between PW-FouCast and traditional unimodal models widens as the lead time increases. This sustained superiority suggests that the explicit frequency-domain alignment of meteorological priors provides a robust and scalable solution to the forecast horizon bottleneck, enabling more reliable long-lead nowcasting.

\vspace{-4pt}
\vspace{-4pt}

\subsection{Case Study}
\begin{figure}[htbp]
  \centering
  \includegraphics[width=0.86\columnwidth]
  {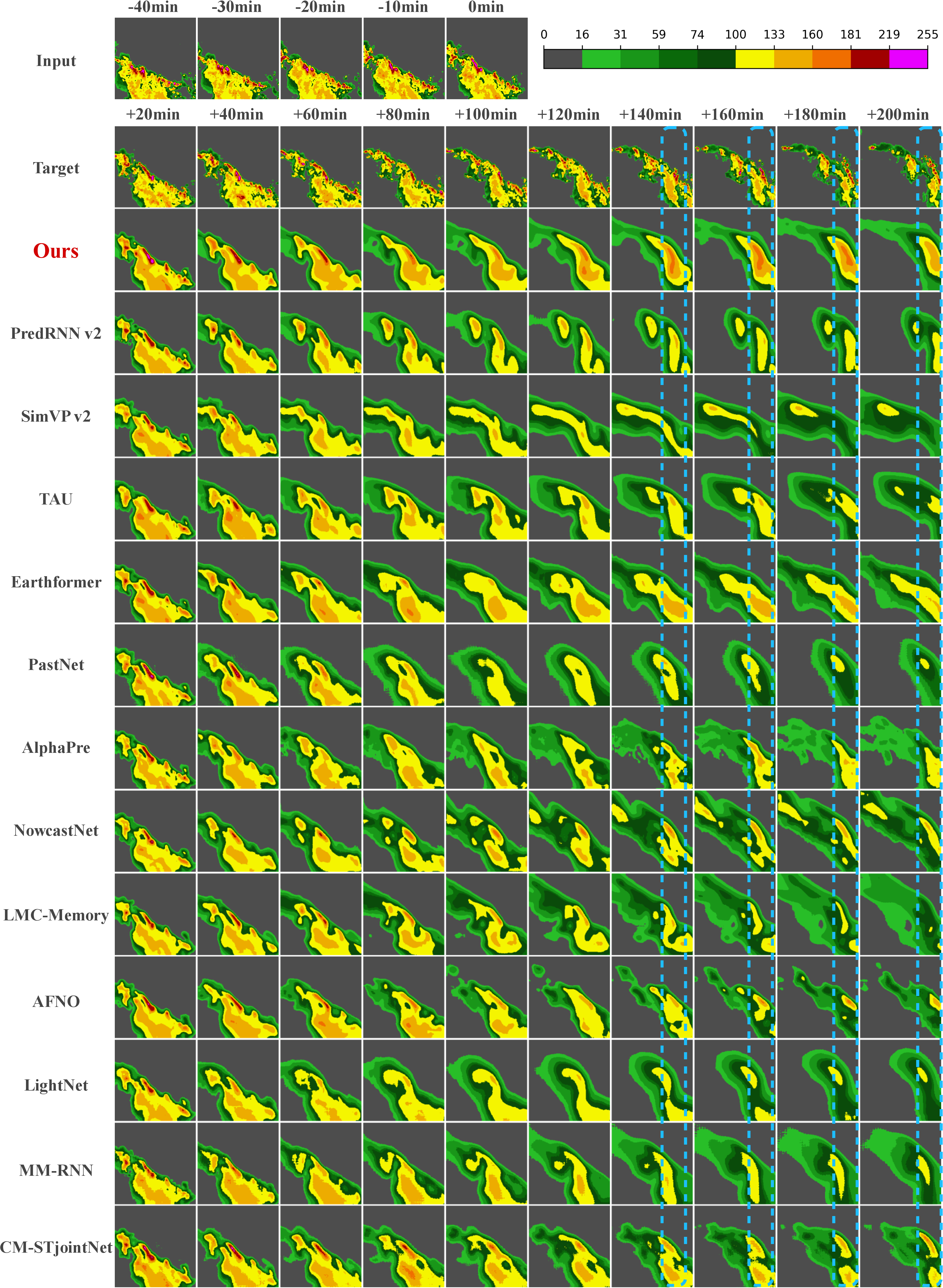}
  \hspace{0.2cm}
  \caption{Qualitative comparison of radar echo predictions on the SEVIR dataset.}
  \label{fig5}
\end{figure}

\begin{figure}[htbp]
  \centering
  \includegraphics[width=0.86\columnwidth]
  {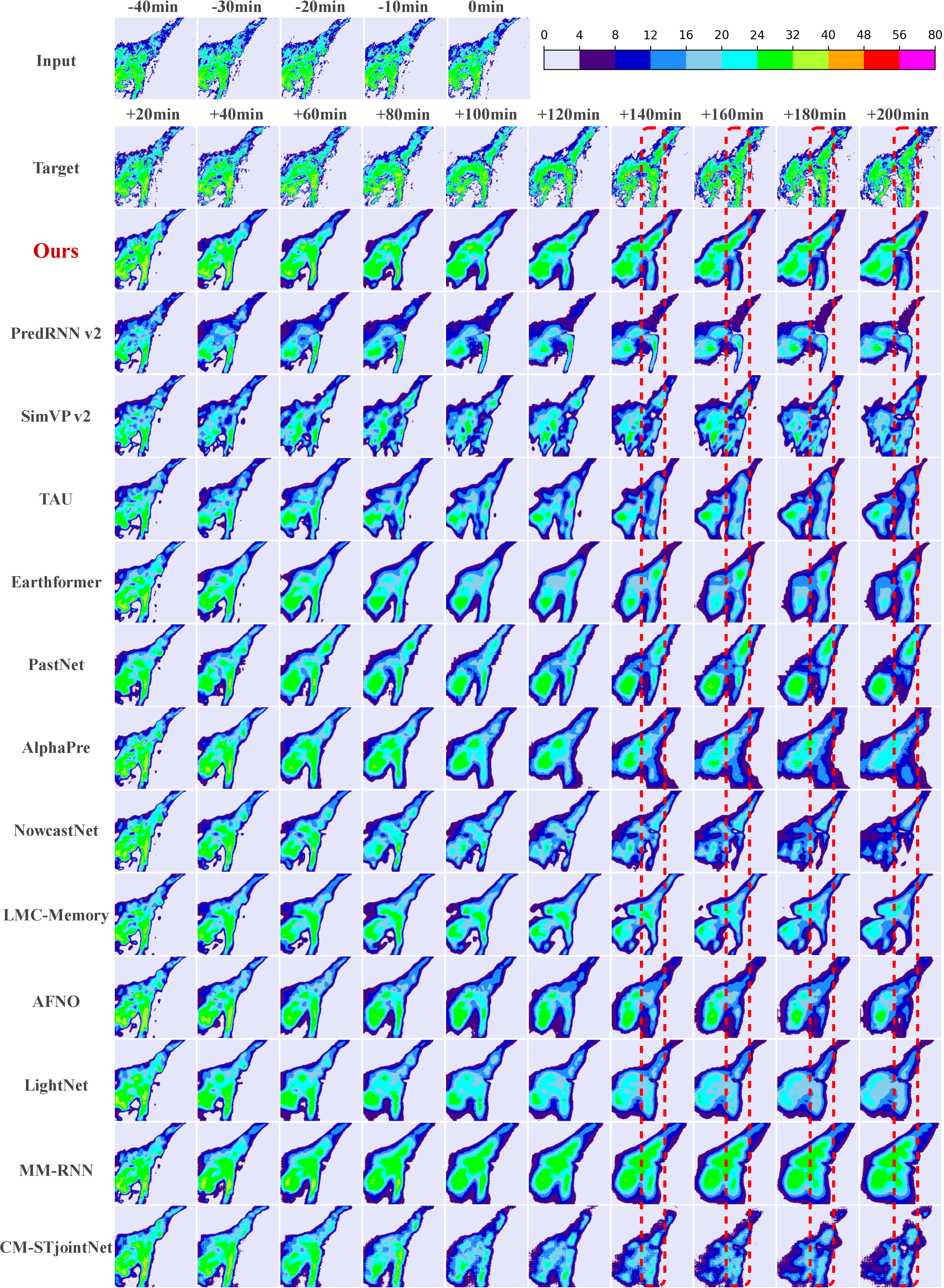}
  \hspace{0.2cm}
  \caption{Qualitative comparison of radar echo predictions on the MeteoNet dataset.}
  \label{fig6}
\end{figure}

Visual evaluation confirms that our model generates physically consistent forecasts with superior structural fidelity. In the SEVIR case study (Fig.~\ref{fig5}), the model accurately captures complex evolution in the precipitation field; notably, it maintains sharp echo structures (blue boxes) even beyond the two-hour mark where baselines typically degrade. Similarly, MeteoNet results (Fig.~\ref{fig6}) show our model maintaining well-defined structures at lead times exceeding 120 minutes (red boxes), outperforming both unimodal and multimodal baselines. This success is driven by the Pangu-Weather-guided frequency modulation and Frequency Memory, which jointly inject large-scale structural priors and preserve fine-grained spatial structures to prevent detail erosion over time.

\subsection{Ablation Study}
The ablation study on the SEVIR dataset (Table~\ref{tab3}) confirms the distinct contributions of each proposed component. PFM enhances radar skill metrics (CSI/HSS) by aligning hidden-layer spectral properties with meteorological priors, while FM primarily reduces spatial errors (MSE/MAE) by correcting phase information using recalled spectral patterns. Their synergy is vital, whereby PFM’s amplitude reweighting and FM’s phase correction jointly sharpen precipitation localization. Finally, IFA recovers high-frequency details typically lost in conventional spectral attention, effectively boosting perceptual scores and ensuring realistic echo boundaries.

\begin{table}[htbp]
  \centering
  \scriptsize
  \setlength{\tabcolsep}{2pt}
  \caption{Ablation study of the proposed modules.}\label{tab3}
{\renewcommand{\arraystretch}{1.1}
  \begin{tabular}{C{0.7cm}|C{0.7cm}|C{0.7cm}|C{0.7cm}|C{0.7cm}|C{0.9cm}|C{0.9cm}|C{0.9cm}|C{0.9cm}}
    \toprule
    \multirow{2}{*}{PFM} & \multirow{2}{*}{FM} & \multirow{2}{*}{IFA} & \multicolumn{1}{c|}{CSI$\uparrow$} & \multicolumn{1}{c|}{HSS$\uparrow$} & \multirow{2}{*}{MSE$\downarrow$} & \multirow{2}{*}{MAE$\downarrow$} & \multirow{2}{*}{PSNR$\uparrow$} & \multirow{2}{*}{SSIM$\uparrow$} \\
    & & & Avg & Avg & & & & \\
    \midrule
    \graycross & \graycross & \graycross & 0.2709 & 0.3630 & 721.6809 & 13.4405 & 23.5702 & 0.5174 \\
    \ding{52} & \graycross & \graycross & 0.2753 & 0.3687 & 697.6037 & 12.9826 & 24.0104 & 0.5593 \\
    \graycross & \ding{52} & \graycross & 0.2704 & 0.3596 & 697.7378 & 13.2158 & 23.9647 & 0.5531 \\
    \ding{52} & \ding{52} & \graycross & {\bfseries 0.2804} & {\bfseries 0.3768} & {\underline{691.1524}} & {\underline{12.6581}} & {\underline{24.0601}} & {\underline{0.5703}} \\
    \ding{52} & \ding{52} & \ding{52} & {\underline{0.2797}} & {\underline{0.3757}} & {\bfseries 676.6416} & {\bfseries 12.5787} & {\bfseries 24.1511} & {\bfseries 0.5789} \\
    \bottomrule
  \end{tabular}
  }
\end{table}

\section{Conclusion}\label{sec:conclusion}
In this paper, we proposed PW-FouCast, a frequency-domain fusion framework for multimodal precipitation nowcasting that overcomes the forecast horizon bottleneck by integrating Pangu-Weather meteorological priors. The architecture leverages three core modules. The Pangu-Weather-guided Frequency Modulation aligns hidden-layer spectral properties with meteorological priors to capture shared structural patterns. The Frequency Memory module employs a learned repository of ground-truth spectral patterns to correct phase information, preserving structural dynamics like expansion and contraction. Finally, Inverted Frequency Attention recovers high-frequency details lost in standard spectral operators through a residual-reinjection mechanism. Achieving SOTA results on SEVIR and MeteoNet, our model demonstrates that phase-aware spectral fusion of foundation model priors effectively enhances accuracy and structural fidelity, providing a robust solution for time-critical meteorological applications.  Future work will investigate the integration of additional observational modalities, such as satellite imagery, to further bolster nowcasting performance.

\section*{Acknowledgment}
This work was supported in part by the Talent Fund of Beijing Jiaotong University under Grant 2025JBRC004, the Foundation of Key Laboratory of Big Data \& Artificial Intelligence in Transportation (Beijing Jiaotong University), Ministry of Education (No. BATLAB202402),  the Foundation of CMA Key Laboratory of Transportation Meteorology (No. JTQX2026M04).


\bibliographystyle{IEEEtran}
\bibliography{references}

\end{document}